%% file: main.tex
\documentclass[10pt,twocolumn]{article} 
\usepackage[utf8]{inputenc}
\usepackage[a4paper,left=2cm,right=2cm,top=2.5cm,bottom=2.5cm]{geometry}
\usepackage{graphicx}
\usepackage{color,soul}
\usepackage{soul}
\usepackage{amsmath}
\usepackage[subrefformat=parens,labelformat=parens]{subcaption}
\usepackage{makecell}
\usepackage{fancyhdr}
\usepackage{amsmath}
\usepackage{amssymb}
\usepackage{tabu}
\usepackage{hyperref}

\begin{document}

\title{Texture Synthesis with Recurrent Variational Auto-Encoder}

\author{Rohan Chandra \quad Sachin Grover \quad Kyungjun Lee\quad Moustafa Meshry \quad Ahmed Taha\\
		\{rohan,kjlee,mmeshry,ahmdtaha\}@cs.umd.edu \quad saching@umd.edu
}
\date{}
\maketitle
\begin{abstract}
We propose a recurrent variational auto-encoder for texture synthesis. A \textit{novel} loss function, FLTBNK, is used for training the texture synthesizer. It is rotational and partially color invariant loss function. Unlike L2 loss, FLTBNK explicitly models the correlation of color intensity between pixels. Our texture synthesizer \footnote{Code available at \href{https://github.com/MoustafaMeshry/draw/}{Github}} generates neighboring tiles to expand a sample texture and is evaluated using various texture patterns from Describable Textures Dataset (DTD). We perform both quantitative and qualitative experiments with various loss functions to evaluate the performance of our proposed loss function (FLTBNK) --- a mini-human subject study is used for the qualitative evaluation.
\end{abstract}

\section{Introduction}
\input{intro.tex}

\section{Background}
\label{sec:background}
\input{background.tex}

\section{Approach}
\label{sec:approach}
\input{approach.tex}

\section{Experimental Results}
\label{sec:result}
\input{result.tex}

\section{Related Works}
\label{sec:relatedworks}
\input{relatedworks.tex}

\section{Conclusion}
\label{sec:conclusion}
\input{conclusion.tex}


\bibliography{ref}
\bibliographystyle{ieeetr}

\end{document}

%% file: intro.tex
\begin{figure*}[!ht]
	\centering
	\includegraphics[scale=0.5]{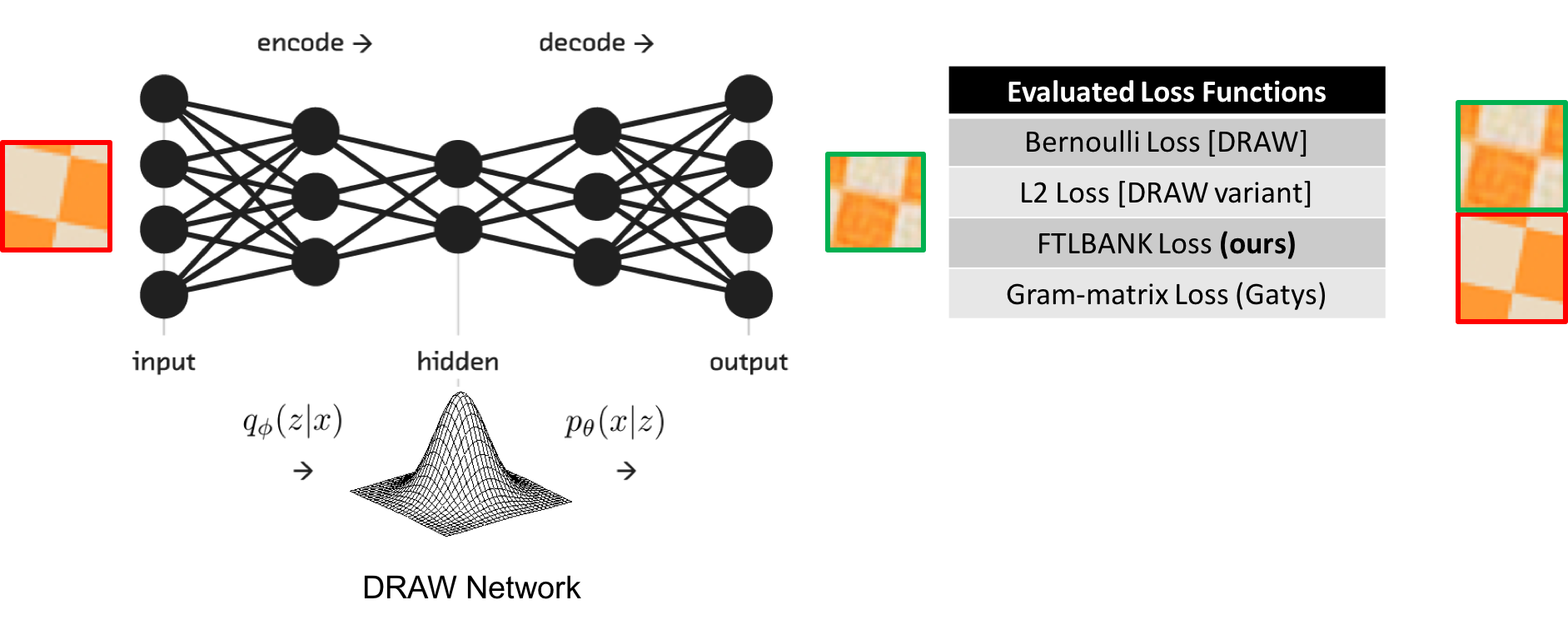}
	\caption{Network Pipeline. In the training phase, a texture and the corresponding neighbor tiles are sampled from DTD. Using center tile as input, DRAW network learns the neighbor tiles. Multiple loss functions are evaluated. In deployment phase, the center tiles is fed as input, and the generated neighbor tile is stitched to the input tile. To expand further, previously generated tiles are fed in DRAW as input.}
	\label{fig:pipeline}
\end{figure*}

Texture synthesis is the process of generating a new texture given a texture sample. Mapping a texture onto surface, scene rendering, occlusion fill-in, lossy image, video compression, and foreground removal are applications for texture synthesis. In this project, DRAW network ~\cite{gregor2015draw}, initially proposed to generate MNIST dataset, is amended to generate textures. Generated tiles are constrained to both having the same texture and aligning smoothly with neighboring tiles. Our texture synthesis model can expand a sample texture and generate a new sample with user-defined dimensions. Figure~\ref{fig:pipeline} shows our pipeline.

We train DRAW to synthesize a texture and enforce smooth alignment between neighboring tiles. We propose a \textit{novel} loss function, FLTBNKs, for training a generative texture network. It is evaluated against $L2$ loss, as a baseline, and the texture loss proposed by Gatys et al.~\cite{gatys2015texture}. 
As we sample texture tiles from DTD, DRAW, a recurrent variational auto-encoder, learns neighboring tiles. Multiple loss functions are evaluated for texture synthesis. In the deployment phase, the trained network generates the four neighboring tiles for a center tile --- an initial sample texture. The generated tiles act as input in the next step to further expand the texture size.

%% file: background.tex
In this section, we describe the frameworks and the loss functions utilized by our project.

\subsection{Deep Recurrent Attentive Writer (DRAW)}
\label{draw}
\begin{figure}[!ht]
	\centering
	\includegraphics[width=0.45\textwidth]{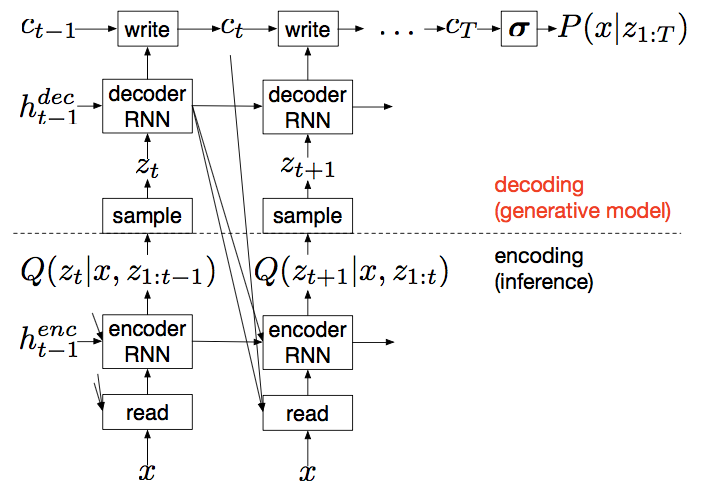}
    \caption{DRAW, a recurrent variational auto-encoder network.}
    \label{fig:draw-architecture}
\end{figure}

Deep Recurrent Attentive Writer \cite{gregor2015draw} (DRAW), introduced by Google DeepMind, is a generative recurrent variational auto-encoder. Figure~\ref{fig:draw-architecture} shows DRAW network architecture. An encoder network determines a distribution over latent variables to capture input data salient information; a decoder network receives samples from the latent distribution and uses them to condition its own distribution over images. This operation is performed iteratively to update attention mechanism that selects "where to read", "where to write", and "what to write".

\subsection{Texture Loss Functions}
We evaluate four loss functions for texture synthesis. Traditional image loss functions such as L2 loss and cross-entropy are considered; yet, other specific texture loss functions are evaluated as well. Sections ~\ref{sec:loss:fb} and ~\ref{sec:loss:vgg} describe filter bank loss and VGG loss -- a recent loss function used for texture synthesis. The equations for these loss functions are summarized in table~\ref{tbl:loss_fns} 

\subsubsection{Filter Bank}
\label{sec:loss:fb}
\begin{figure}[ht]
			\centering
			\includegraphics[scale=0.5]{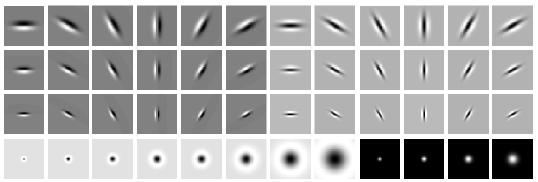}
        \caption{LM filters bank}
        \label{fig:flt_bnk}
 \end{figure}

Varma et al.~\cite{varma2005statistical} propose a statistical approach for texture classification. Before deep neural networks made their mark, it was a very competitive classification approach. Despite being primitive, it is popular for its simplicity and accuracy. To generate a texture descriptor, a filter bank is applied on a texture image. The Leung-Malik (LM), shown in figure ~\ref{fig:flt_bnk}, is a commonly used filter bank. After filtering, a texture image becomes a L-dimensional image, where L is the number of filters. Each pixel, L-dimensional vector, gets classified to pre-trained cluster centers called textons. Using these textons, a texture image is represented by a texton histogram. We propose such concept for training our texture synthesis network.

\subsubsection{VGG}
\label{sec:loss:vgg}
Another recent loss function for texture generation is very deep convolutional networks (VGG). It is first introduced in a large-scale image classification work~\cite{simonyan2014very}. This texture loss function is proposed by Gatys at el. ~\cite{gatys2015texture}  for texture generation and adapted later for image style transfer. Basically, feature maps generated at different VGG-network layers, by similar textures, have high correlation. Thus, the correlation between input texture $x$ and the synthesized image $\hat x$ is a quantitative metric. To elaborate, $x$ and $\hat x$ are fed into the VGG-network independently, each producing their own set of feature maps, $F^l \in \mathbb{R}^{N_l \textrm{x} M_l}$ and $\hat F^l\in \mathbb{R}^{N_l \textrm{x} M_l}$, where each layer $l$ has $N_l$ distinct feature maps each of size $M_l$ when vectorized. $F_{jk}^l$ is the activation of the $j^{th}$ filter at position $k$ in layer $l$. The correlation between these feature maps  are stored in a matrix $G$ and $\hat G$ respectively. $G^l \in \mathbb{R}^{{N_l \textrm{x} N_l}}$ is defined as:

\[G_{ij}^l = \sum_k F_{ik}^l F_{jk}^l\]  
	
	The distance, between two textures, is the weighted sum of layer-wise distance.
	
	\[D_l = \dfrac{1}{4N_l^2 M_l^2} \sum_{i,j} (G_{ij}^l - \hat G_{ij}^l)^2\]
	\[\mathcal{D}^{total}(x,\hat x) = \sum_{l=0}^L w_l D_l\]


%% file: approach.tex
\begin{figure}[!ht]
	\centering
	\includegraphics[width=1.0\linewidth]{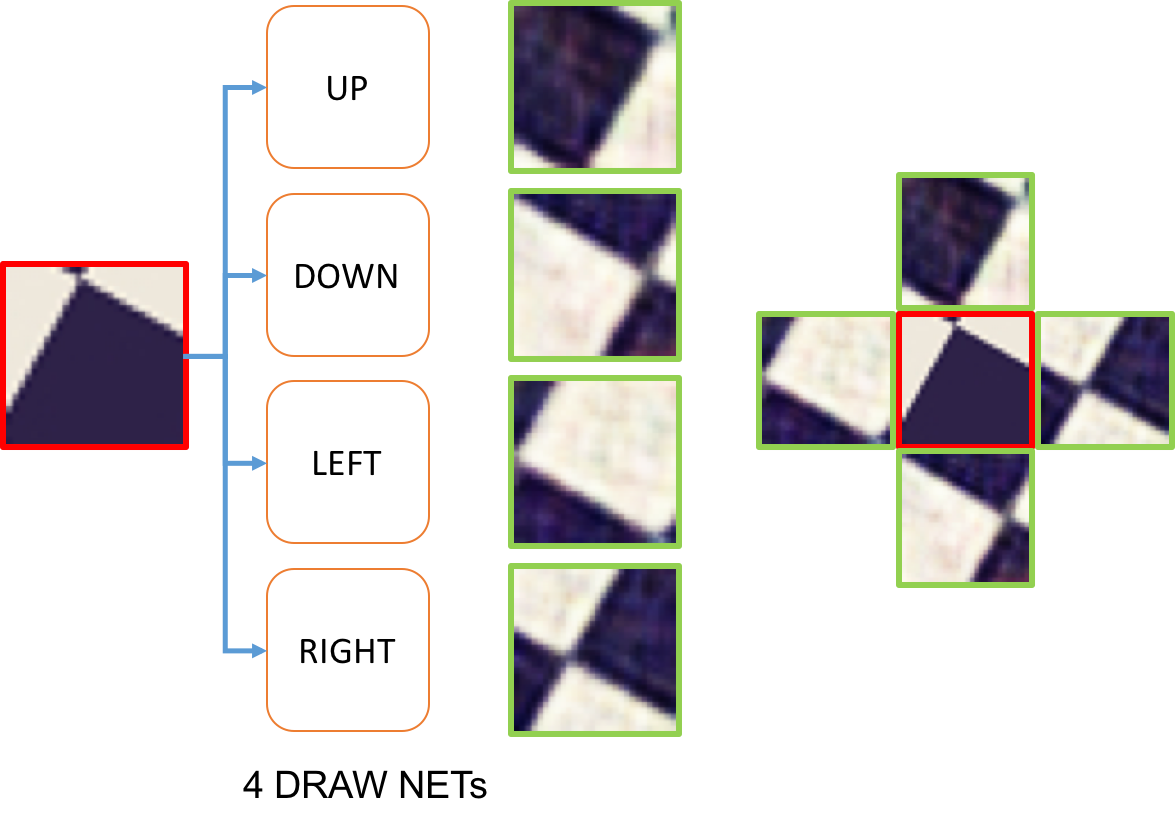}
	\caption{Four DRAW networks are trained independently to generate the four neighboring tiles. Generated tiles are stitched to the input center tile to output the final texture}
	\label{fig:approach}
\end{figure}

The DRAW \cite{gregor2015draw} network is a variational generative model to generate images that cannot be distinguished from real data with the naked eye. Our project extends the application of the DRAW network to generate neighboring textures tiles from the DTD dataset~\cite{cimpoi14describing}. Our approach is illustrated in figure~\ref{fig:approach}, and table ~\ref{tbl:loss_fns} shows all the reconstruction loss functions used for evaluation. 
The following subsections describe our approach details

\begin{table}
\centering
\begin{tabular}{|c|c|}
\hline 
\multicolumn{2}{|c|}{Evaluated Reconstruction Loss Functions}  \\
\hline 
\hline 
Cross Entropy & $y\log { \hat { y }  }  + (1-y)\log { (1-\hat {y })  }  $ \\ 
\hline 
L2 &  $(y-\hat { y } )^2$ \\ 
\hline 
FB & $(LM(y)-LM(\hat { y }))^2$ \\ 
\hline 
VGG &  \makecell{$ (1/(4N_l^2 M_l^2)) \sum_{i,j} (G_{ij}^l - \hat G_{ij}^l)^2$ \\ $G_{ij}^l = \sum_k F_{ik}^l F_{jk}^l$} \\ 
\hline 
\end{tabular} 
\caption{Reconstruction loss functions evaluated for texture synthesis. In our experiments, different combinations are evaluated. Total network loss is reconstruction loss plus latent loss.}
\label{tbl:loss_fns}
\end{table}

\subsection{Texture Generation}
To train our network, five regular texture patterns are selected from DTD. During each training epoch, tiles are sampled from these five texture patterns. The DRAW network is fed the center tile and evaluated using both the latent and reconstruction loss. Different combination of loss functions, shown in table~\ref{tbl:loss_fns}, evaluates the reconstructed texture tile $L^c$. At the same time, a latent loss $L^z$ evaluates the latent variable distribution.\\
We use a Gaussian latent variable distribution for two reasons. First, the gradient of a function of the samples with respect to the distribution parameters can be easily obtained using the so-called reparameterization trick~\cite{rezende2014stochastic,kingma2015auto}. This makes it straightforward to back-propagate unbiased, low variance stochastic gradients of the loss function through the latent distribution. Since our latent prior is a standard Gaussian with mean zero and standard deviation one. The Kullback-Leibler divergence is computed by the closed form equation:
\begin{align}
{ L }^{ z }=\sum _{ t=1 }^{ T }{ KL(Q({ Z }_{ t }|{ h }_{ t }^{ enc })||P({ Z }_{ t })) } \\
{ L }^{ z }=\frac { 1 }{ 2 } \left( \sum _{ t=1 }^{ T }{ { \mu  }_{ t }^{ 2 }+{ \sigma  }_{ t }^{ 2 }-\log { { \sigma  }_{ t }^{ 2 } }  }  \right) -\frac { T }{ 2 } 
\end{align}

The original filter bank, proposed in ~\cite{varma2005statistical}, computes a histogram for a texture image. Histograms are not differentiable; so, back-propagation becomes infeasible. To overcome such obstacle, loss is computed directly from LM filter responses. Thus no clustering assignment or binning is performed. Throughout this report, FB refers to LM filter bank response; while FTLBNK refers to a combination of FB with other losses defined in section~\ref{sec:regularization}. The total loss $L$ for the network is the sum of the reconstruction and latent losses $L = L^c + L^z$. Once the four networks are trained, neighboring tiles are generated by feeding center tile and stitching the tiles accordingly. 

\subsection{FLTBNK Loss Function}
\label{sec:regularization}
We propose a novel loss function, called FLTBNK. It utilize the idea that filter banks are rotationally and partially color invariant.  The filter banks normalize the images and their filter responses. Thus, it is partially invariant to changes in illumination intensity. Neighboring pixels contribute to a pixel filter's response. Thus, unlike L2 loss, a pixel intensity is correlated with its neighbors. Such characteristics nominate FLTBNK loss function to train our texture generative network. 

Texture ~\ref{fig:regularization}\subref{fig:reg_fb} shows the noisy reconstructed texture image when using FB only. To address the color problem, we apply a color regularizer that optimize for the mean color within a tile; this generated texture~\ref{fig:regularization}\subref{fig:reg_fb_color}. In texture ~\ref{fig:regularization}\subref{fig:reg_fb_tv}, we reduce the noise by adding total variation loss to FB. To eliminate both color and noise problems, both total variation and color regularization are added to FB loss as shown in texture~\ref{fig:regularization}\subref{fig:reg_fb_color_tv}.
\input{figures_tex/fb_regularization}

%% file: figures_tex/fb_regularization.tex
\begin{figure}[!ht]
	\begin{subfigure}[b]{0.14\textwidth}
      \centering
	  \includegraphics[scale=2.5]{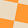}
      \caption{Original}
      \label{fig:reg_org}
	\end{subfigure}
	~
	\begin{subfigure}[b]{0.14\textwidth}
      \centering
	  \includegraphics[scale=2.5]{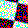}
      \caption{FB} 
      \label{fig:reg_fb}
	\end{subfigure}
	~
	\begin{subfigure}[b]{0.14\textwidth}
      \centering
	  \includegraphics[scale=2.5]{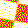}
      \caption{FB+col reg}
      \label{fig:reg_fb_color}
	\end{subfigure}
	~
	\begin{subfigure}[b]{0.14\textwidth}
      \centering
	  \includegraphics[scale=2.5]{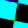}
      \caption{FB+TV}
      \label{fig:reg_fb_tv}
	\end{subfigure}
	~
	\begin{subfigure}[b]{0.18\textwidth}
      \centering
	  \includegraphics[scale=2.5]{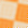}
      \caption{FB+ col reg +TV}
      \label{fig:reg_fb_color_tv}
	\end{subfigure}
	\caption{Effect of different regularizations on the filter-bank loss. First row shows (a) input textures, (b)filter-bank loss with no regularization, and (c) filter-bank loss with a color regularizer respectively. Second row shows (d)filter-bank loss with total variation and (e) filter-bank loss, total variation with a color regularizer.}
	\label{fig:regularization}
  \end{figure}

%% file: result.tex





In this section, we show qualitative results of our model on two tasks:  texture reconstruction and large texture generation from a small texture sample. Then, a small quantitative evaluation of the synthesized texture is provided. Finally, we show a human subject study to evaluate generated textures.

\input{figures_tex/reconstruction.tex}

\subsection{Input Texture Reconstruction}
The first evaluation metric is sample texture reconstruction.
Figure~\ref{fig:reconstruction} shows reconstructed textures using four different losses: cross-entropy, L2, FTLBNK and VGG. It shows that corss-entropy, L2 or FTLBNK generates a blurry output. This happens because DRAW is a variational auto-encoder where the loss function is a variational upper bound on the log-likelihood of the data.

\begin{figure*}[t!]
	\centering
	\includegraphics[scale=0.25]{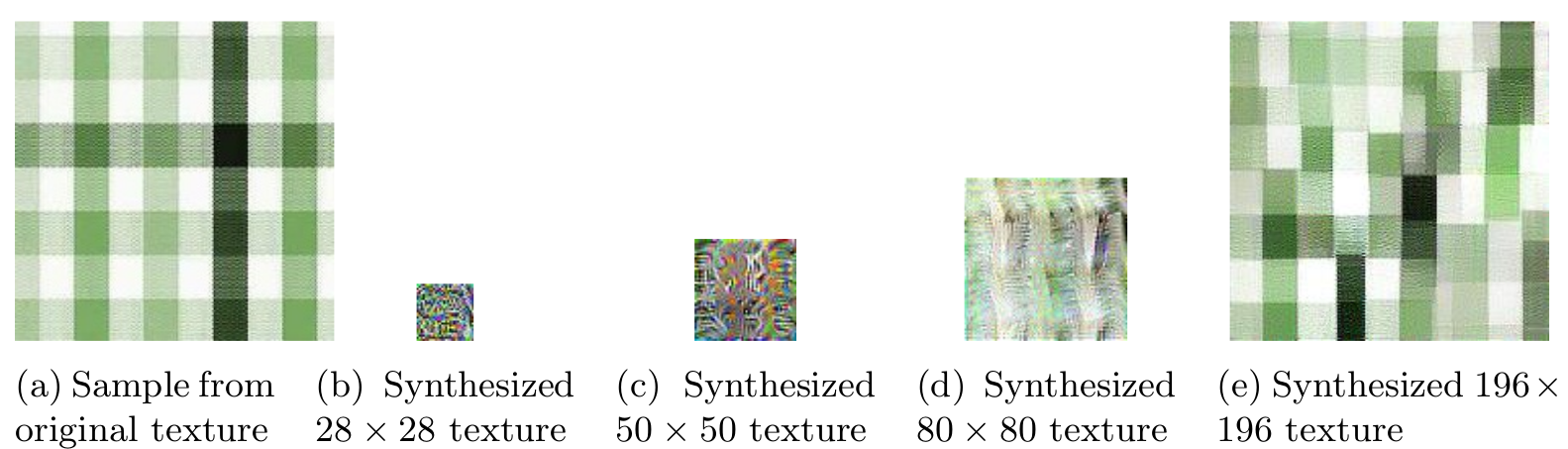}

	\caption{Texture synthesis using an implementation of Gatys \emph{et al.}~\cite{gatys2015texture}. Results show that VGG loss needs large input images to work properly.}
	\label{fig:gatys}
\end{figure*}

All the losses evaluated optimize the generated data likelihood using a latent representation $P_\theta(\tilde{X}|Z)$. In some sense we are averaging out and we are trying to fit a unimodal distribution to a multimodal solution which results in blurry images. On the other hand, the VGG
loss didn't work well with our approach. One reason is that the VGG loss requires large input images, since the VGG architecture contains many pooling layers.
To verify this reason, we ran a public implementation of ~\cite{gatys2015texture}, and
we confirmed that it only works with large images, while it generates almost random noise with small inputs (such as 28x28), as shown in figure~\ref{fig:gatys}

\subsection{Texture Generation}
\input{figures_tex/generation.tex}

The second evaluation metric is sample texture expansion. Figure~\ref{fig:generation} shows qualitative output of generated textures using two different losses: L2 and FTLBNK.
Generation using the L2 loss captures texture colors well, but it doesn't perform very well on more complicated textures. For example, the first sample of green texture contains many fine details, and 
the last sample texture contains curves instead of a simple chess-board tiles. In both cases, the L2 loss performs poorly. On the other hand, FTLBNK does a better job capturing the textures outlines, but it suffers from a color-problem.
This is probably because the color regularization term from section~\ref{sec:regularization} focuses largely on the mean color value for the red, green and blue channels independently and less on the correlation between different color channels.\\
To numerically evaluate generated textured without bias, generated texture is compared with original textures using two distance metric: VGG and texture histogram as originally proposed in ~\cite{varma2005statistical}.  

\begin{figure}[h!]
\centering
\begin{subfigure}[b]{0.40\textwidth}
			\centering
			\includegraphics[scale=0.5]{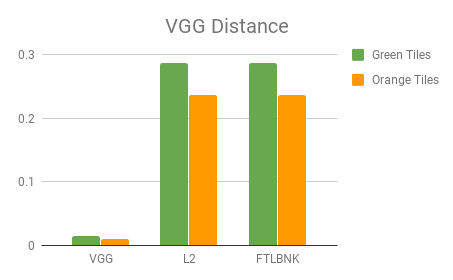}
		\end{subfigure}
		\begin{subfigure}[b]{0.40\textwidth}
			\centering
			\includegraphics[scale=0.5]{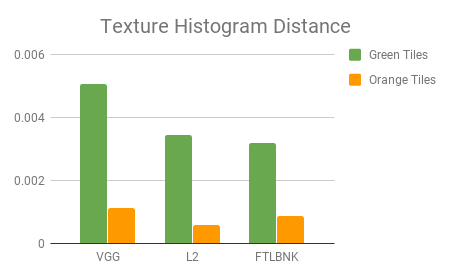}
		\end{subfigure}
        \caption{Distance between original texture and synthesized texture using both VGG and histogram distances. The orange and green textures from figure ~\ref{fig:reconstruction} are used for evaluation. }
        \label{fig:texture_distance}
\end{figure}

Figure ~\ref{fig:texture_distance} shows the VGG and histogram distance between original and synthesized texture. The orange and green textures from figure ~\ref{fig:reconstruction}, first and third columns, are used for evaluation. From the figure, we conclude that a network trained with VGG loss generates a texture with smaller VGG distance. On the other hand, a network trained with FB loss generates a texture with smaller histogram distance. A merit for histogram distance is using textons which are clusters trained one a texture dataset. Thus, we believe the histogram distance is more informative than the VGG distances based on solely a pair of images information. 

\subsection{Human Subjects Evaluation}
To further evaluate the loss functions used to generated textures, we conducted a short human subject study. The study was designed as a survey asking human subjects to rate the quality of resemblance (5 point Likert Scale) between the original and generated textures for each loss function. All participants are machine learning students, thus, our study suffers some level of selection bias. Due to time constraints, we obtained 18 responses, but this is enough to run a statistical test. To avoid confirmation bias, the loss functions names are hidden in the survey. As the independent variable is categorical, and the dependent variable is ordinal (5 point Likert Scale), we perform the Kruskal-Wallis Test (non-parametric). The p-value is $<< 0.05$ and therefore we reject the null hypothesis that the medians of all groups (loss functions) are equal. 

Figure~\ref{fig:responses} shows a summary of the responses. The key observations is that although traditional losses (cross-entropy and L2) have more `Excellent' ratings, our proposed loss, FLTBNK, seems to get better overall ratings. FLTBNK gets many `Very Good' ratings compared to other losses. We hypothesize that the results of traditional losses have high variance, while results of FLTBNK are more consistent, and so, we see that a majority of the responses to traditional losses are `Acceptable', whereas for FLTBNK the majority of responses are `Very Good'.
\input{figures_tex/responses.tex}    

%% file: figures_tex/reconstruction.tex
\begin{figure*}[!ht]
	\centering
	\includegraphics[scale=0.3]{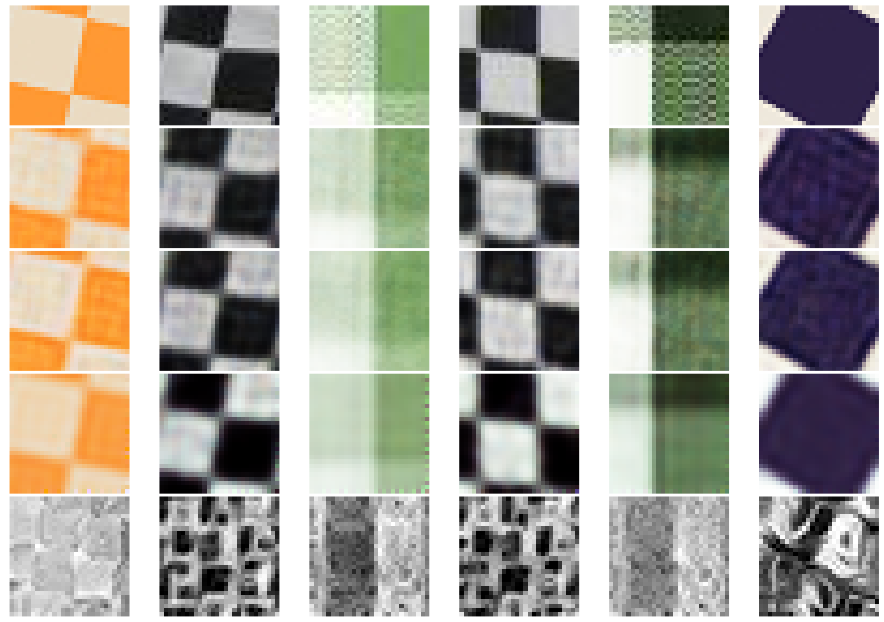}

	\caption{Sample texture reconstruction. First row shows the original input texture. The following rows show reconstructed textures using the cross-entropy, L2, FTLBNK and VGG losses respectively. Each tile is 28x28 pixels.}
	\label{fig:reconstruction}
  \end{figure*}

%% file: figures_tex/generation.tex
\begin{figure*}[!ht]
	\centering
	\begin{subfigure}[b]{0.13\textwidth}
	  \includegraphics[scale=0.35]{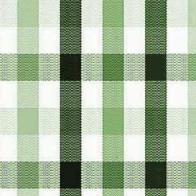}
	\end{subfigure}
	~
	\begin{subfigure}[b]{0.13\textwidth}
	  \includegraphics[scale=0.35]{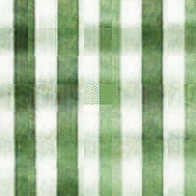}
	\end{subfigure}
	~
	\begin{subfigure}[b]{0.13\textwidth}
	  \includegraphics[scale=0.35]{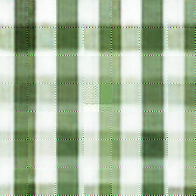}
	\end{subfigure}
~
	\begin{subfigure}[b]{0.13\textwidth}
	  \includegraphics[scale=0.35]{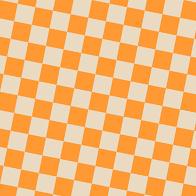}
	\end{subfigure}
	~
	\begin{subfigure}[b]{0.13\textwidth}
	  \includegraphics[scale=0.35]{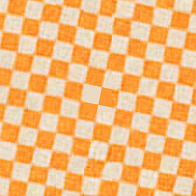}
	\end{subfigure}
	~
	\begin{subfigure}[b]{0.13\textwidth}
	  \includegraphics[scale=0.35]{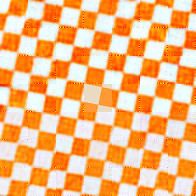}
	\end{subfigure}

	\begin{subfigure}[b]{0.13\textwidth}
	  \includegraphics[scale=0.35]{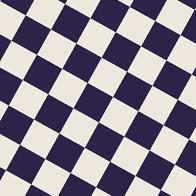}
	\end{subfigure}
	~
	\begin{subfigure}[b]{0.13\textwidth}
	  \includegraphics[scale=0.35]{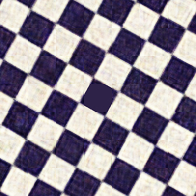}
	\end{subfigure}
	~
	\begin{subfigure}[b]{0.13\textwidth}
	  \includegraphics[scale=0.35]{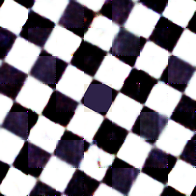}
	\end{subfigure}
~
	\begin{subfigure}[b]{0.13\textwidth}
	  \includegraphics[scale=0.35]{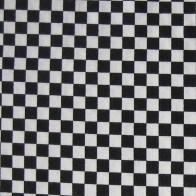}
	\end{subfigure}
	~
	\begin{subfigure}[b]{0.13\textwidth}
	  \includegraphics[scale=0.35]{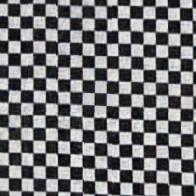}
	\end{subfigure}
	~
	\begin{subfigure}[b]{0.13\textwidth}
	  \includegraphics[scale=0.35]{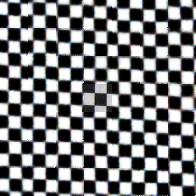}
	\end{subfigure}

	\begin{subfigure}[b]{0.13\textwidth}
	  \includegraphics[scale=0.35]{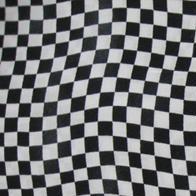}
	\end{subfigure}
	~
	\begin{subfigure}[b]{0.13\textwidth}
	  \includegraphics[scale=0.35]{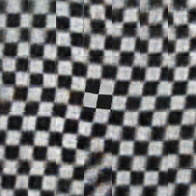}
	\end{subfigure}
	~
	\begin{subfigure}[b]{0.13\textwidth}
	  \includegraphics[scale=0.35]{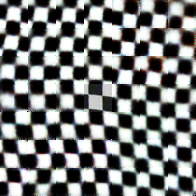}
	\end{subfigure}

	\caption{Generating texture. Input is only a 28x28 center tile, while the output size is 196x196. For each image,
    the left: the original texture; middle: generated texture using L2 loss; right: generated texture using filter-bank loss}
	\label{fig:generation}
  \end{figure*}

%% file: figures_tex/responses.tex
\begin{figure*}[!h]
	\centering
	\includegraphics[width=\textwidth]{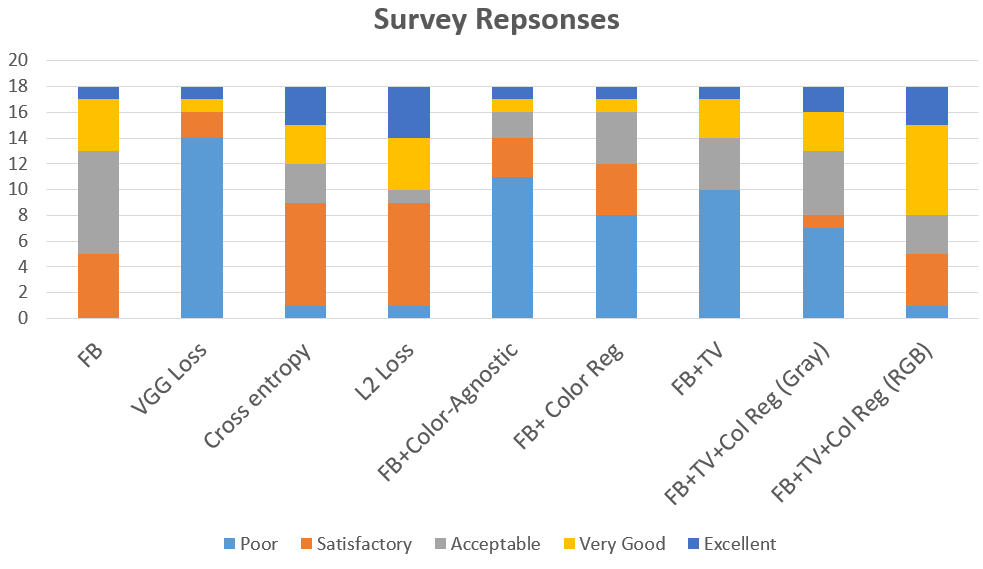}
	\caption{Ratings from 18 participants on a 5-point scale. Participants prefer traditional losses, like cross entropy and L2, outputs in terms of excellent rating. Yet, FLTBNK loss, last column, gets better overall ratings. FLTBNK gets many `Very Good' ratings compared to other losses.}
	\label{fig:responses}
  \end{figure*}

%% file: relatedworks.tex
Several recent work uses convolutional neural networks for texture synthesis. Gatys et al. ~\cite{gatys2015texture} proposed a new texture synthesis model based on the feature spaces of convolutional neural networks. This model computes the Gram matrices on feature maps to transform textures into a feature space. However, they do not evaluate how suitable the Gram matrices are to recognize textures, which might lead to low accuracy of the model on certain textures. Also, this model can only generate a textures of same size as input texture, while our model can expand a textures using a small sample. Yet, it is able to generate more challenging textures than our model. 

While Gatys et al. model requires a lot of memory to process the inputs, a new model proposed by Ulyanov et al.~\cite{ulyanov2016texture} alleviate the memory overheads. The new model reduces memory overhead during training by putting computational burden to a learning stage, while maintaining the same accuracy. Yet, this memory-less model does not expand a small texture sample, which our model does.

To avoid deep neural network inherently high computational costs, one research work on texture synthesis introduced computational cost reduction using Markovian generative adversarial networks~\cite{li2016precomputed}. This model utilizes adversarial generative networks in a Markovian setting to capture the feature statistics. As this model improves the performance and is able to generate continuous images with an image of a small fixed size, it would be interesting to compare our model with this model in terms of the synthesized image quality. Evaluation between our proposed model and the Markovian generative adversarial model is a worth-noting extension missing due to time constraints.

%% file: conclusion.tex
We use DRAW, a variational generative model, for texture synthesize. Our model expands a small sample texture by generating neighboring tiles. A novel texture loss function, FLTBNK, is proposed to impose color correlation between generated pixels. A qualitative study, with 18 human subjects, shows that a combination of FLTBNK, variational loss and a color regularize is competitive as l2 loss. 
While exact color matching is attractive attribute for human subjects, it is weakly required in texture context. Thus, FLTBNK, a combination of filter bank response, color variational loss and a color regularizer is more intuitive for texture generation.

